\documentclass[letterpaper, 10 pt, conference]{ieeeconf}  

\IEEEoverridecommandlockouts 

\usepackage{hyperref}
\usepackage{amssymb} 
\usepackage{amsmath,amssymb,amsfonts}
\usepackage{algorithmic}
\usepackage{graphicx}
\usepackage{textcomp}
\usepackage{xcolor,makecell}
\usepackage{dblfnote}
\usepackage{booktabs}
\let\labelindent\relax
\usepackage{enumitem}
\usepackage{array}
\usepackage{multirow,tabularx}
\usepackage{tcolorbox}
\usepackage[table]{xcolor}

\colorlet{unseenyellow}{yellow!20}

\title{\LARGE \bf
3D CAVLA: Leveraging Depth and 3D Context to \\ Generalize Vision Language Action Models for Unseen Tasks
}

\author{Vineet Bhat$^1$, Yu-Hsiang Lan$^2$, Prashanth Krishnamurthy$^1$, Ramesh Karri$^1$, and Farshad Khorrami$^1$
\\ \\
$^1$New York University Tandon School of Engineering \\ $^2$New York University Courant Institute of Mathematical Sciences 
}

\begin{document}

\maketitle
\thispagestyle{empty}
\pagestyle{empty}


\begin{abstract}

Robotic manipulation in 3D requires effective computation of $N$ degree-of-freedom joint-space trajectories that enable precise and robust control. To achieve this, robots must integrate semantic understanding with visual perception to transform real-world observations into low-level control for object interaction. Recent advances in Vision-Language-Action (VLA) models have shown promise by mapping RGB images and language instructions to task space velocities, typically trained on large datasets of teleoperated demonstrations. However, these models often struggle with generalization beyond their training distributions. In this work, we introduce \textbf{3D-CAVLA}, a novel finetuning framework that enhances task generalization of VLA policies by incorporating three key components: (i) chain-of-thought reasoning for structured decision-making, (ii) depth-aware perception for 3D spatial understanding, and (iii) task-oriented region-of-interest detection for focused manipulation. Extensive experiments in the LIBERO simulation environment demonstrate that 3D-CAVLA achieves an average success rate of 98.1\% across diverse in-domain task suites. On unseen tasks, 3D-CAVLA delivers an absolute improvement of 8.8\% in success rate, underscoring the benefits of 3D scene awareness for robust generalization. We validate our approach on real-world tabletop experiments demonstrating that the proposed model translates effectively from simulation to physical robots. 3D-CAVLA achieves over a 3$\times$ faster training convergence and delivers a 25\% gain in success rate on unseen real world tasks. We will release our code and datasets to foster community-driven research and accelerate progress in VLAs for robotic manipulation.

\end{abstract}


\section{INTRODUCTION}

The ability to perceive the environment, respond dynamically, and manipulate objects is a central challenge in robotics. Humans exhibit this capability effortlessly, developed through years of experiential learning that integrates perception, reasoning, and motor skills to handle familiar and novel scenarios. Replicating such adaptability into robots, however, is  difficult. Advances in foundation models, particularly in vision and language understanding, have shown progress toward bridging this gap. Vision-Language Models (VLMs), such as ChatGPT, leverage large-scale pre-training on internet data to interpret visual inputs, comprehend natural language, and generate contextually relevant outputs. These capabilities have been applied in visual question answering~\cite{c65}, visual grounding~\cite{c1}, and task planning~\cite{c42}, skills which are important to robotics.

Vision-Language-Action (VLA) models extend VLMs by predicting robot actions instead of textual responses~\cite{c3,c4} which are used to actuate the joints of a robot to perform a task. When pre-trained on real-world datasets and fine-tuned on high-quality teleoperated demonstrations, VLAs achieve strong performance ($\ge$95\%) on in-distribution tasks such as ``scoop pretzels into a bowl,'' but their generalization to unseen tasks remains limited. Recent work has shown that incorporating reasoning within VLA training objectives can improve out-of-domain performance~\cite{c55,c66,c67}. 



In this work, we propose targeted improvements to strengthen spatial and contextual reasoning in VLAs. Our framework, \textbf{3D-CAVLA} (\textbf{3D} \textbf{C}ontext-\textbf{A}ware \textbf{V}ision-\textbf{L}anguage \textbf{A}ction), incorporates three core innovations: (i) Chain-Of-Thought (CoT) style narrative prompts that enrich task context and promote structured reasoning, (ii) depth-aware features derived from workspace point clouds that provide 3D spatial understanding, and (iii) task-oriented Region-Of-Interest (ROI) pooling that focuses attention on visually relevant regions. 3D-CAVLA is a finetuning paradigm and can be applied after pre-training the vision-language encoders of any standard VLA. Our experiments show that 3D-CAVLA improves in-distribution performance and generalization to unseen tasks. We benchmarked 3D-CAVLA against state-of-the-art VLAs in the LIBERO simulation environment and validated it on robot tabletop experiments. Our main contributions are:
\begin{enumerate}
    \item A finetuning framework for existing VLAs that integrates chain-of-thought prompting, depth-based 3D features, and region-of-interest pooling, improving spatial perception, task reasoning, and LIBERO in-distribution success to 98.1\%.
    \item Extensive zero-shot evaluation on unseen tasks showing an 8.8\% absolute improvement over existing VLA baselines. 3D-CAVLA achieves a 3x faster training convergence while operating at the standard 4.3Hz frequency policy rollout of modern VLAs.
    \item Real-robot experiments under seen, similar, and unseen settings evaluate performance under increasing difficulty, with 3D-CAVLA achieving a 25\% improvement over the nearest baseline on unseen tasks.
\end{enumerate}


\section{RELATED WORKS}
\label{sec:related-works}

\noindent \textbf{Foundational Models in Robotics. } LLMs can generate high-level robotic execution plans based on task inputs and environmental context ~\cite{c43}. However, a recurring challenge with LLMs is their tendency to hallucinate, generating plans that are not physically feasible ~\cite{c39}. To enhance robustness, LLMs require real-world grounding, which can be achieved through feedback from the environment ~\cite{c42}, integration with visual perception systems~~\cite{c59}, or human-in-the-loop interventions such as question-answering ~\cite{c65}. VLMs, trained on vast image-text datasets, excel at visual reasoning tasks ~~\cite{c1} and have been applied to a range of robotics grounding problems such as encoding 3D semantic memory~\cite{c12}, pose estimation~\cite{c73}, language guided object manipulation~\cite{c41,c69}, and robotic navigation~\cite{c15}. 

\noindent \textbf{Vision-Language Action Models.} VLMs pretrained on internet-scale real-world data possess a vast knowledge base. By replacing the text prediction head with robot action heads, these models can be further finetuned with robot demonstration data, mapping vision-language inputs to robot task space velocities~\cite{c3}. Prior work has trained VLAs on large datasets of robotic demonstration videos and language instructions to predict robot actions for manipulation.  Early VLAs demonstrated strong performance in simulation and single robot manipulation~\cite{c4}. However, many of these models are limited by their closed-source nature or extremely large parameter sizes~\cite{c4}. OpenVLA~\cite{c21}, a
representative open-source autoregressive VLA, stands out as one of the first approaches to release a compute efficient and scalable VLA with a parameter size of $\approx$7B, specifically fine-tuned on robot demonstration data from the Open-X Embodiment corpus~\cite{c33}. OpenVLA-OFT~\cite{c22} further improves inference efficiency and task performance by incorporating parallel decoding, action chunking, continuous action representations, with an L1 regression objective.

\noindent \textbf{Spatial Grounding and Reasoning for VLA Task Generalization.} Recent studies highlight the scaling challenges of directly translating visual frames and language instructions into robot actions, as the volume of task demonstrations increases~\cite{c55}. A prominent direction to enhance task generalization of VLAs involves pre-training multimodal encoders with self-supervision using unlabeled human task demonstrations and diverse video planning datasets~\cite{c52}. This stage aims to learn robust representations without relying on explicit action labels. Some works employ teacher-student RL frameworks to refine action policies~\cite{c45}. Other approaches integrate pre-trained models such as CLIP~\cite{c37}. For instance, CLIP-RT~\cite{c20} uses CLIP’s visual and textual encoders to associate RGB frames with instructions and textual actions during pre-training. Fine-tuning then focuses on selecting actions from a fixed set of classes at each timestep. DynaMo~\cite{c6} adopts a self-supervised strategy, employing both forward and inverse dynamics models to train visual encoders for future observation prediction, enabling robust action forecasting. Recent work has explored integrating proprioception and feedback mechanisms to dynamically correct erroneous actions~\cite{c46}. 

A key limitation of VLAs is their dependency on direct input-output mappings without intermediate reasoning. CoT prompting encourages step-by-step reasoning grounded in language, visual observations, and physical actions. Integrating textual descriptions, keypoints, or subgoal images provides structured guidance for planning and action prediction~\cite{c55}. CoT-VLA~\cite{c63} introduces chain-of-thought by predicting future RGB frames as visual goals, paired with short action sequences to reach them. ECoT~\cite{c55} interleaves high-level planning and subtask decomposition with embodied reasoning steps grounded in object bounding boxes and end-effector positions before action generation, enforcing the VLA to reason before acting. ECoT-Lite~\cite{c67} explores reasoning during pre-training to avoid extra token prediction during inference while matching the rollout frequency of standard VLAs. CoA-VLA~\cite{c66} uses affordances as intermediate predictions with object locations, grasp poses, and waypoint-based motion trajectories before action generation. SpatialVLA~\cite{c64} incorporates 3D positional encodings to enhance spatial awareness in policy learning. 3D-CAVLA integrates chain-of-thought directly into the input language instructions, producing narrative-style task decompositions that strengthen long-horizon reasoning. Task-relevant features are pooled using a region-of-interest mask computed once before inference, while depth encoding is performed in parallel with vision-language encoding. This setup preserves the inference frequency of the VLA without prediction of additional reasoning tokens. Unlike prior approaches that strengthen geometric representations or introduce intermediate reasoning, 3D-CAVLA combines both with explicit geometric grounding and language-level task decomposition. 

Depth perception is valuable for robotic manipulation, as it enhances geometric understanding and spatial reasoning~\cite{c44}. Concurrent to this work, recent VLA research has begun to explicitly incorporate 3D geometric priors and depth-aware representations to improve spatial grounding and generalization. For example, GeoVLA~\cite{c70} processes point-cloud features with a parallel point embedding network fused with semantic features, 
while QDepth-VLA~\cite{c72} injects structural cues through auxiliary quantized depth prediction. 

\begin{figure*}[!ht]
    \vspace{5pt}
    \centering
    \includegraphics[width=1\linewidth]{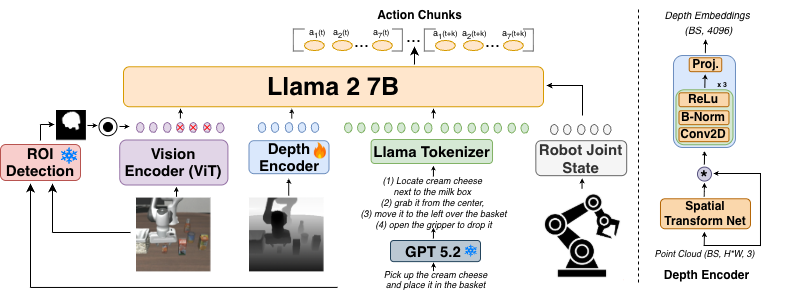}
    \caption{3D-CAVLA integrates chain-of-thought style narrative task descriptions, depth embeddings and region of interest pooling to improve scene awareness of vision-language-action modeling. While GPT5.2 and ROI Detection are frozen components, our depth encoder is a lightweight PointNet~\cite{c35} inspired trainable network (RHS) with spatial invariance transformation, convolution blocks and linear projections. We show only one camera view for brevity, though adding a gripper-mounted view is straightforward as in prior work. Task-aware feature pooling is applied only to the static-camera view. }
    \label{fig:proposed_framework}
    \vspace{-10pt}
\end{figure*}

\section{METHODOLOGY}
\label{sec-method}

Our approach builds upon a popular VLA, OpenVLA-OFT~\cite{c22}, which consist of vision and language encoders for processing world observations, combined with an LLM backend that learns mappings to robot actions. 3D-CAVLA is a systematic improvement in finetuning, rather than a new architecture, and can be incorporated into a wide range of VLA architectures. We first provide a concise primer on the key components of VLAs, and then introduce our proposed finetuning paradigm. 

\subsection{Vision-Language to Action Modeling}

VLA policies map multimodal observations of a robot's environment and internal states to low-level control actions. We denote a task instruction $x^{\text{text}}$, a set of visual observations $x^{\text{vis}}$, and the robot's proprioceptive state $x^{\text{prop}}$. The objective is to learn a policy $\pi_\theta$ parameterized by $\theta$, that predicts an $N$-dimensional vector $a_t \in \mathbb{R}^N$ at each timestep $t$:
\begin{equation}
    a_t = \pi_\theta(x^{\text{vis}}_t, x^{\text{text}}, x^{\text{prop}}_t).
\end{equation}

In this work, we restrict the formulation to a robotic arm, where $a_t$ denotes the end-effector 6D pose and a binary gripper open/close state.

\paragraph{Vision-Language Embeddings}  
Visual inputs $x^{\text{vis}}_t$ consist of RGB frames captured from robot's cameras. A vision encoder $f_{\text{vis}}$ extracts patch-level embeddings. ($v_t$), as 
\begin{equation}
    v_t = f_{\text{vis}}(x^{\text{vis}}_t), \quad v_t \in \mathbb{R}^{d_v \times P}
\end{equation}
where $P$ is the number of image patches and $d_v$ is the embedding dimension. Task instructions $x^{\text{text}}$ are tokenized and processed by a language encoder $f_{\text{text}}$:
\begin{equation}
    l = f_{\text{text}}(x^{\text{text}}), \quad l \in \mathbb{R}^{d_l \times T}
\end{equation}
where $T$ is the number of tokens, $l$ is the language embedding and $d_l$ is the embedding dimension. Vision and language embeddings are projected into a common latent space ($\tilde{v}_t$, $\tilde{l}$) of dimension $d$ via learnable projections $W_v$ and $W_l$:
\begin{equation}
    \tilde{v}_t = W_v v_t, \quad \tilde{l} = W_l l, \quad \tilde{v}_t \in \mathbb{R}^{d \times P}, \tilde{l} \in \mathbb{R}^{d \times T}.
\end{equation}

\paragraph{Proprioception Embeddings ($p_t$)}  
The robot's proprioceptive state $x^{\text{prop}}_t \in \mathbb{R}^{d_p}$ (e.g., joint angles and gripper state) is encoded by a multilayer perceptron (MLP):
\begin{equation}
    p_t = f_{\text{prop}}(x^{\text{prop}}_t), \quad p_t \in \mathbb{R}^d.
\end{equation}

\paragraph{Fusion and Policy Learning}  
The fused representation at timestep $t$ is obtained by concatenating:
\begin{equation}
    z_t = \text{Concat}(\tilde{v}_t, \tilde{l}, p_t), \quad z_t \in \mathbb{R}^{d \times L}, \quad L=P+T+1
\end{equation}
where $L$ denotes the total sequence length after fusion. This sequence $z_t$ is passed into an LLM backbone $f_{\text{LLM}}$:
\begin{equation}
    h_t = f_{\text{LLM}}(z_t), \quad h_t \in \mathbb{R}^d.
\end{equation}

A linear head $W_a$ maps LLM output to final action vector:
\begin{equation}
    a_t = W_a h_t, \quad a_t \in \mathbb{R}^N.
\end{equation}

\paragraph{Training Objective}  
The model is trained on teleoperated demonstrations 
$\mathcal{D} = \{(x^{\text{vis}}_{1:T}, x^{\text{text}}, x^{\text{prop}}_{1:T}, a_{1:T})\}$ 
where each trajectory has length $T$. The objective is a regression loss over predicted joint velocities:  
\begin{equation}
    \mathcal{L}(\theta) = \frac{1}{|\mathcal{D}|} \sum_{(x,a) \in \mathcal{D}} 
    \frac{1}{T} \sum_{t=1}^{T} 
    \left\| \pi_\theta(x^{\text{vis}}_t, x^{\text{text}}, x^{\text{prop}}_t) - a_t \right\|_1
\end{equation}
where $\|\cdot\|_1$ denotes the $\ell_1$ loss, commonly used for learning continuous representations of actions.


\begin{figure}[!ht]
  \centering
  \small
  \begin{tcolorbox}[title=Chain Of Thought Prompting with GPT 5.2]
    You are in the command of a robot manipulator to complete a task involving various objects in scene. Your job is to break down the given instruction into smaller steps based on real-world intuition and scene camera view to ensure precise object grasping and placement\\
    
    Some examples are given below: \\

    \textbf{Task Instruction:} Put both pots on the stove \\
    \textbf{Steps:} Grasp first pot, place on stove leaving some space, grasp second pot, place on stove next to first pot. \\
    
    \{more examples\}\\
    
    Use the information above to create step-by-step plan for given task instruction. Remember to only use the given objects and standard grasping, moving and placement actions that can be achieved with a parallel gripper.\\ 
    
    \textbf{Task Instruction:} \{task\_input\}\\
    \textbf{RGB view:} \{rgb\_input\}\\
    \textbf{Steps:}
  \end{tcolorbox}
  
  \caption{LLM prompt to decompose task instructions into executable steps that can be generalized across seen and unseen tasks.}
  \vspace{-10pt}
  \label{fig:cot-prompt}
\end{figure}

\begin{figure}[!ht]
    \vspace{5pt}
    \centering
    \includegraphics[width=1\linewidth]{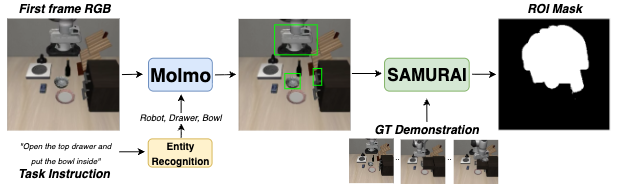}
    \caption{Task-Aware ROI detection pipeline. Task-relevant entities are identified via named-entity recognition, localized with object detection, and tracked across frames. The resulting binary mask (right) pools visual features, guiding motion to task-relevant regions and enhancing vision-language fusion for improved generalization. By focusing perception on critical regions, TA-ROI helps policies remain robust on unseen tasks.}
    \label{fig:tarp-fig}
    \vspace{-10pt}
\end{figure}

\subsection{Our Approach: 3D-CAVLA}

To improve generalization beyond seen tasks, we propose \textbf{3D-CAVLA}, a systematic VLA improvement (see Fig.~\ref{fig:proposed_framework})  that augments the baseline modeling framework with three key components: (i) chain-of-thought narrative instructions for enriched contextual reasoning, (ii) depth-based point cloud embeddings for 3D spatial awareness, and (iii) task-aware region of interest pooling for focused perception. 

\noindent \textbf{Chain-of-Thought Narrative Instructions.}  
Task instructions $x^{\text{text}}$ (Eq.~2) are typically provided as plain natural language. However, these single sentence instructions may not capture the structured reasoning required for generalization. To address this, we augment instructions with CoT decompositions $\tilde{x}^{\text{text}}$, where each task is rewritten as a sequence of intermediate steps. For instance, the instruction “Grab the ball and place it in the basket” is reformulated as: “Locate ball $\rightarrow$ grasp at center $\rightarrow$ move over basket $\rightarrow$ release.” During inference, unseen tasks (e.g., “Move the orange into the basket”) benefit from this decomposition since only the first step (locating the unseen object) changes, while the subsequent reasoning structure remains reusable. We generate $\tilde{x}^{\text{text}}$ automatically using GPT-5.2 (see Fig.~\ref{fig:cot-prompt}), and encode them using the LLM's tokenizer $f_{\text{text}}$ (Eq.~3) to produce enriched text embeddings $\tilde{l}$. These embeddings enhance temporal coherence and guide the LLM backbone toward structured action generation.  Unlike learning reasoning tokens within the policy~\cite{c55,c67}, a frozen LLM for CoT reasoning prevents overfitting to training distributions and preserves generalization for unseen tasks.

\noindent \textbf{Integrating Depth Features.}  
Prior VLAs rely on RGB images $x^{\text{vis}}$. Robust manipulation requires reasoning over object geometry and spatial layout. Modern sensors provide RGB-D input, from which we recover metric 3D point clouds $P$. Given a depth map $D \in \mathbb{R}^{B\times H\times W}$ and camera intrinsics $(f_x,f_y,c_x,c_y)$, each pixel $(h,w)$ in batch $b$ is back projected:
\[
\begin{aligned}
Z_{b,h,w} &= D_{b,h,w},\\[2pt]
X_{b,h,w} &= \frac{\,U_{h,w}-c_x\,}{f_x}\,Z_{b,h,w},\\[2pt]
Y_{b,h,w} &= \frac{\,V_{h,w}-c_y\,}{f_y}\,Z_{b,h,w}.
\end{aligned}
\]
Stacking $(X,Y,Z)$ yields $P \in \mathbb{R}^{B \times H \times W \times 3}$. A lightweight PointNet-inspired encoder $f_{\text{depth}}$ transforms $P$ into embeddings $d_t \in \mathbb{R}^d$, which are concatenated with $\tilde{v}_t$, $\tilde{l}$, and $p_t$ (Eqs.~4–5) before fusion (Eq.~6). Since $f_{\text{depth}}$ is compact ($\sim$1M parameters), we deploy separate encoders per camera view without prohibitive cost. This integration provides fine-grained spatial cues crucial for manipulation of novel objects. 

\noindent \textbf{Task-Aware Region of Interest Detection (TA-ROI).}  
Visual embeddings $\tilde{v}_t$ encode all patches in an image, yet many patches are irrelevant to the manipulation objective. To focus on task-relevant regions, we estimate ROI masks $M \in \{0,1\}^{H\times W}$ during training. Given task instruction $x^{\text{text}}$, we apply named-entity recognition~\cite{c54} to extract objects and locations, detect them with Molmo~\cite{c7}, and track their motion with SAMURAI~\cite{c51} (see Fig. \ref{fig:tarp-fig}). Resulting trajectories define a binary mask $M$ used to pool visual features:
\[
\tilde{v}_t^{\text{ROI}} = \text{Pool}(\tilde{v}_t, M).
\]
This concentrates the fused representation $z_t$ (Eq.~6) on regions critical for action generation. To prevent over-dependence on masks and preserve contextual cues, we stochastically disable ROI pooling for 30\% of training samples.  This improves zero-shot generalization while marginally reducing in-distribution accuracy (Table~\ref{tab:ablations}). During inference when we do not have a sample tele-operated task for object tracking, we use Molmo~\cite{c7} and SAM 2~\cite{c68} to generate relevant object and gripper masks and compute a union to capture the task relevant image region.

\noindent \textbf{Discussion on modularity.} 3D-CAVLA integrates two frozen modules for CoT reasoning and task relevant ROI along with a lightweight trainable depth encoder within a standard VLA formulation. This modular design avoids retraining larger reasoning models and the need to annotate reasoning traces within robot datasets. It also provides flexibility, since stronger external models can be substituted without changing or re-training the policy. The tradeoff is an increased dependency on external components and a complex system pipeline compared to an end-to-end model. In 3D-CAVLA, CoT reasoning and ROI masks are computed once before rollout, and hence they do not affect real-time robot control frequency. The depth encoder adds only a lightweight trainable branch that runs in parallel with the vision and language encoders of the VLA, resulting in a negligible impact on rollout speed (4.4 Hz without depth vs. 4.3 Hz with depth). We view 3D-CAVLA as a practical systems-level finetuning approach for studying how modular reasoning and 3D perception affects task generalization in VLAs.

\noindent \textbf{Experimental Setup.}  
All experiments are conducted on a single NVIDIA A100 GPU with batch size 8. Chain-of-thought decompositions and ROI masks are pre-computed offline for efficiency. We adopt LoRA fine-tuning along with action chunking~\cite{c22}, similar to other baselines.

\begin{table*}[!ht]
\vspace{10pt}
\caption{Results on the LIBERO Benchmark. 3D-CAVLA shows consistent improvement across all task suites in the dual camera setup. Most baselines overfit to the tasks and thus the margins are quite narrow. We utilize open-sourced code repositories of each of the baselines to train and test in LIBERO. All scores are reported in success rate ($\%$). Results for OTTER are taken from the original paper; since it does not report results on Libero-Long, we mark both Libero-Long and the average as "-".}
  \centering
  \setlength{\tabcolsep}{15pt}
  \begin{tabular}{lccccc}
    \toprule
    \multicolumn{6}{c}{\textbf{Policy Setup: Single stationary third person camera + Language Instruction}} \\
    \midrule
    & Spatial & Object & Goal & Long & Average \\
    \midrule
    Diffusion Policy~\cite{c5}
    & 78.3 & 92.5 & 68.3 & 50.5 & 72.4 \\
    Octo~\cite{c60}                             & 78.9 & 85.7 & 84.6 & 51.1 & 75.1 \\
    Diffusion Transformers~\cite{c61}  & 84.2 & \textbf{96.3} & 85.4 & 63.8 & 82.4 \\
    OpenVLA~\cite{c21}  & 84.7 & 88.4 & 79.2 & 53.7 & 76.5 \\
    OTTER~\cite{c16}  & 84.0 & 89.0 & 82.0 & - & - \\
    CoA-VLA~\cite{c66} & 85.3 & 93.1 & \textbf{85.8} & 55.0 & 79.8 \\
    \textbf{Ours: 3D-CAVLA} & \textbf{86.1} & 94.7 & 82.9 & \textbf{66.8} & \textbf{82.6} \\
    \midrule\midrule
    \multicolumn{6}{c}{\textbf{Policy Setup: Third person camera + Wrist camera + Robot states + Language Instruction}} \\
    \midrule
    & Spatial & Object & Goal & Long & Average \\
    \midrule
    Multimodal Diffusion Transformer~\cite{c61}  & 78.5 & 87.5 & 73.5 & 64.8 & 76.1 \\
    $\pi_{0}$~\cite{c62} & 96.8 & 98.8 & 95.8 & 85.2 & 94.2 \\
    OpenVLA‑OFT~\cite{c22}  & 97.6 & 98.4 & 97.9 & 94.5 & 97.1 \\
    \textbf{Ours: 3D-CAVLA} & \textbf{98.2} & \textbf{99.8} & \textbf{98.2} & \textbf{96.1} & \textbf{98.1} \\
    \bottomrule

  \end{tabular}
  \vspace{-10pt}
  \label{tab:libero-benchmark-results}
\end{table*}

\section{RESULTS}
\label{sec-results}

We evaluate our method in three stages. First, we compare against established VLA baselines on the LIBERO benchmark under in-distribution settings. Second, we assess generalization by designing a set of unseen tasks to test zero-shot capabilities. Finally, we validate deployment in the real world with tabletop experiments on the Franka arm. All policies are evaluated in an open-loop execution setting, where actions are predicted autoregressively without additional feedback beyond RGB-D observations.

\subsection{Comparisons on LIBERO Benchmarks}

LIBERO benchmark has four task suites, each probing a different capability of a policy:  
\begin{enumerate}[leftmargin=*]
    \item \textbf{LIBERO-Spatial:}  
    Tasks that manipulate the \emph{same} object but require placing it in different locations.
    
    \textit{Example: Pick up the black bowl between the plate and the ramekin and place it on the plate.}

    \item \textbf{LIBERO-Object:}  
    Tasks with a fixed target location but a \emph{different} object to manipulate.  
    
    \textit{Example: Pick up the milk and place it in the basket.}

    \item \textbf{LIBERO-Goal:}  
    Tasks in which the robot must achieve a higher-level goal beyond simple pick-and-place.  
    
    \textit{Example: Open the top drawer and put the bowl inside.}

    \item \textbf{LIBERO-Long:}  
    Long-horizon tasks that sequentially manipulate multiple objects, testing extended reasoning. 
    
    \textit{Example: Turn on the stove and put the moka pot on it.}
\end{enumerate}

Following prior work~\cite{c21,c22}, we fine-tune independently on every suite, using 50 demonstrations per task and evaluating over 50 trials. Results are reported in Table~\ref{tab:libero-benchmark-results}.  

With only third-person RGB input and text instructions, 3D-CAVLA improves success rates on the Spatial and Long suites and slightly outperforms the Diffusion Transformer~\cite{c61} on average. Chain-of-thought instructions provide reusable semantic structure across related tasks, resulting in more precise action generation. Adding robot proprioception and an additional camera further improves performance (last four rows in Table~\ref{tab:libero-benchmark-results}). With the inclusion of depth maps, 3D-CAVLA consistently surpasses all baselines across the four suites, demonstrating the value of augmenting 2D inputs with 3D spatial cues. Depth features prove particularly beneficial in crowded environments, where accurate localization of target objects is critical. Our findings highlight the benefits of richer spatial representations for manipulation.  

\begin{table}[!ht]
  \centering
  \caption{Success-rate (in $\%$) of Diffusion Policy (DP), OpenVLA-OFT (OFT), ECoT* and 3D-CAVLA (3DC) on 10 unseen tasks. 3D-CAVLA decomposes unseen tasks into seen steps and applies TA-ROI, enabling better task generalization.}
  \renewcommand{\arraystretch}{1.00}
  \setlength{\tabcolsep}{2.8pt}
  \begin{tabular}{m{5cm}cccc}
    \toprule
    \textbf{Task Instruction} & \textbf{DP} & \textbf{OFT} & \textbf{ECoT*} & \textbf{3DC} \\
    \midrule
    Place the white and yellow mug on the plate & 22 & 32 & 48 & \textbf{60} \\
    \midrule
    Put the ketchup on top of the cabinet & 56 & 74 & 78 & \textbf{82}\\
    \midrule
    Pick up the chocolate pudding at the back and put it in the top drawer of the cabinet & 38 & \textbf{58} & 56 & 52 \\
    \midrule
    Stack the right bowl on the left bowl and put the chocolate pudding in the tray & \textbf{6} & 0 & 0 & 0\\
    \midrule
    Put the chocolate pudding on the plate & 60 & 78 & \textbf{80} & \textbf{80} \\
    \midrule
    Place the cream cheese and soup inside the basket & 48 & 66 & 72 & \textbf{74} \\
    \midrule
    Grab the white bowl and keep it on the stove & 6 & \textbf{12} & \textbf{12} & 10 \\
    \midrule
    Grab the chocolate pudding and place it on the bowl. Then place both items on the tray & 4 & 6 & 10 & \textbf{24} \\
    \midrule
    Turn on the stove and put the bowl on it & 12 & 14 & 22 & \textbf{38}\\
    \midrule
    Place mug inside right compartment of the caddy & 18 & 24 & 28 & \textbf{32}\\
    \midrule
    \textbf{Average} & 27.0& 36.4 & 40.6 & \textbf{45.2} \\
    \bottomrule
  \end{tabular}
  \label{tab:libero-unseen-results}
  \vspace{-10pt}
\end{table}

\newcolumntype{C}[1]{>{\centering\arraybackslash}m{#1}}

\begin{table*}[ht]
\vspace{10pt}
\caption{Qualitative comparisons on unseen LIBERO tasks. We show first, middle, and last frames of each inference. The final two rows depict failures where both models misidentify target object or get distracted by seen objects.}
\centering
\small
\setlength{\tabcolsep}{6pt}
\begin{tabular}{C{3.8cm}  C{6.2cm}  C{6.2cm}}
\toprule[1pt]
\textbf{Task Instruction} & \textbf{OpenVLA-OFT} & \textbf{3D-CAVLA}\\
\midrule[0.5pt]
Put the chocolate pudding on the plate &
  \includegraphics[width=\linewidth]{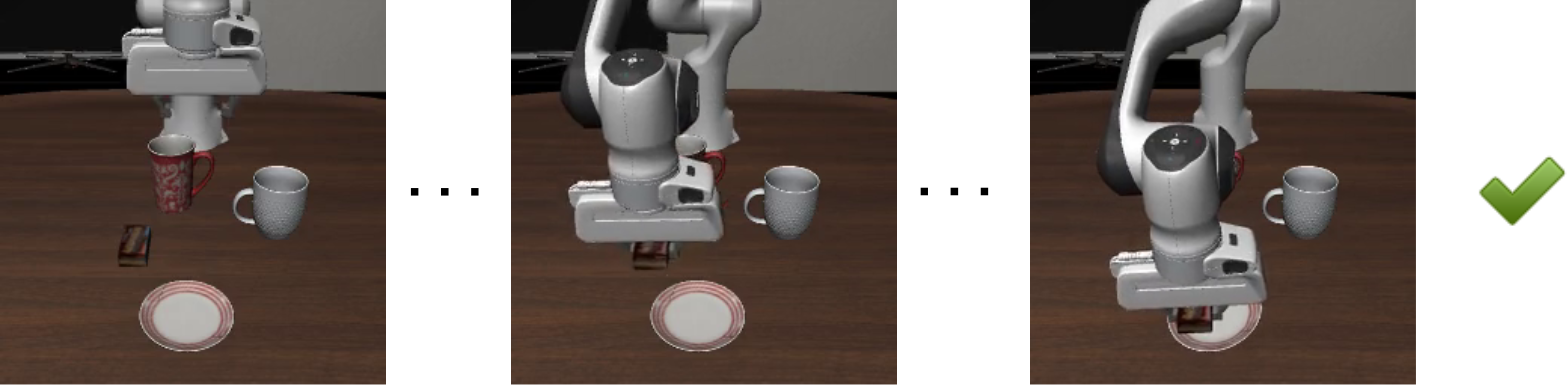} &
  \includegraphics[width=\linewidth]{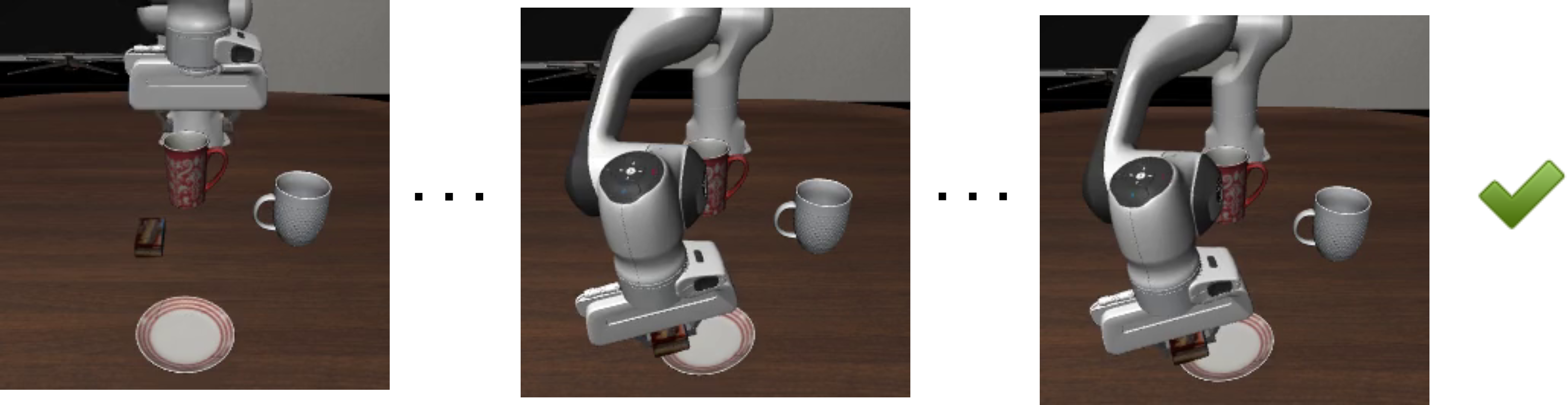}\\
\midrule
Place the white and yellow mug on the plate &
  \includegraphics[width=\linewidth]{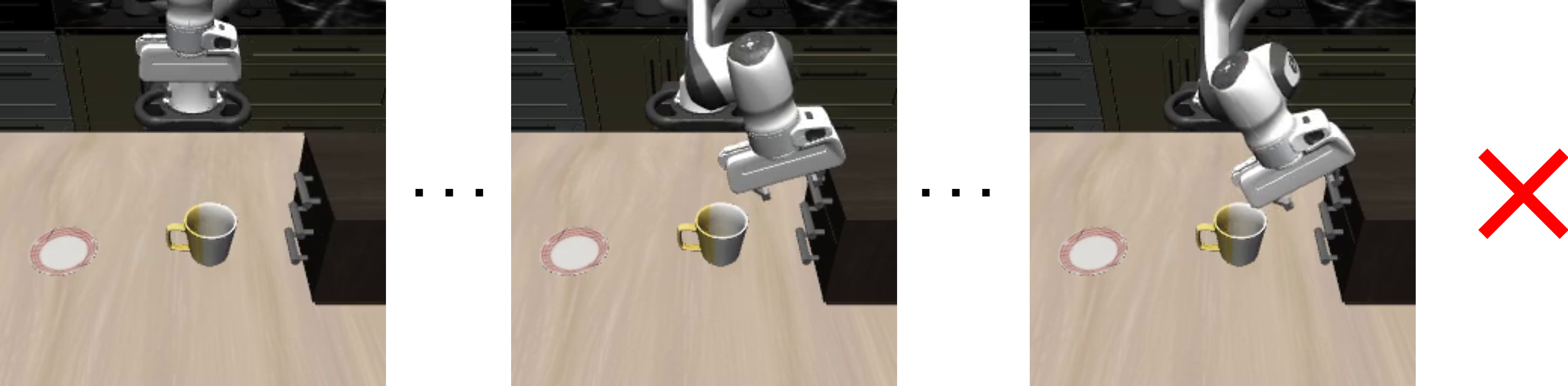} &
  \includegraphics[width=\linewidth]{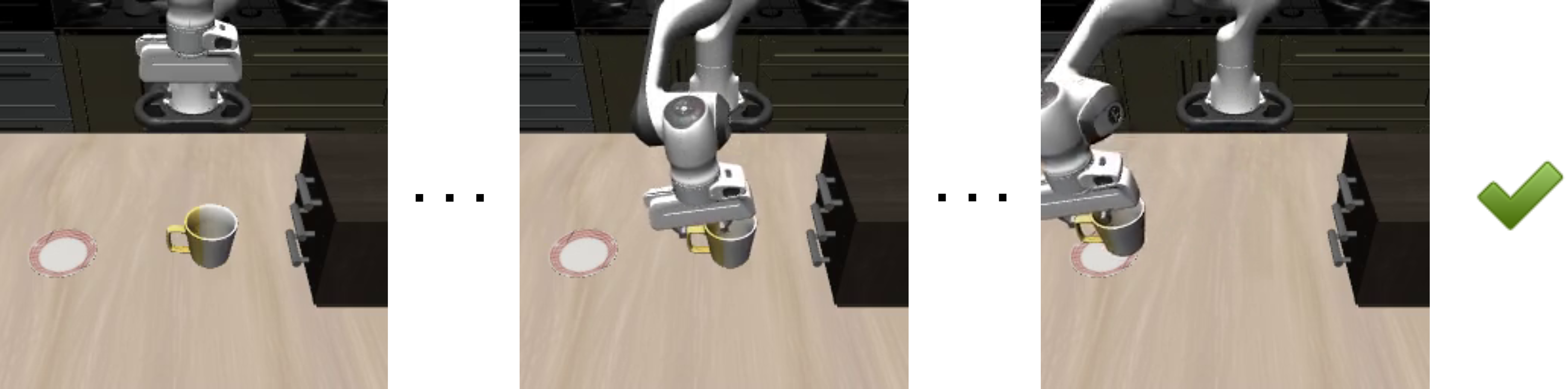}\\
\midrule
Turn on the stove and put the bowl on it &
  \includegraphics[width=\linewidth]{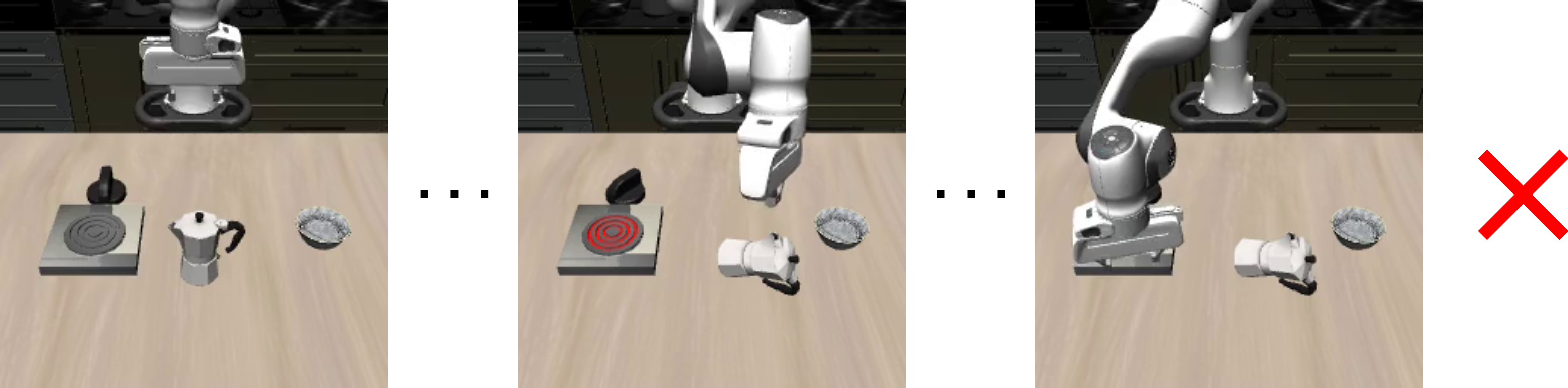}&
  \includegraphics[width=\linewidth]{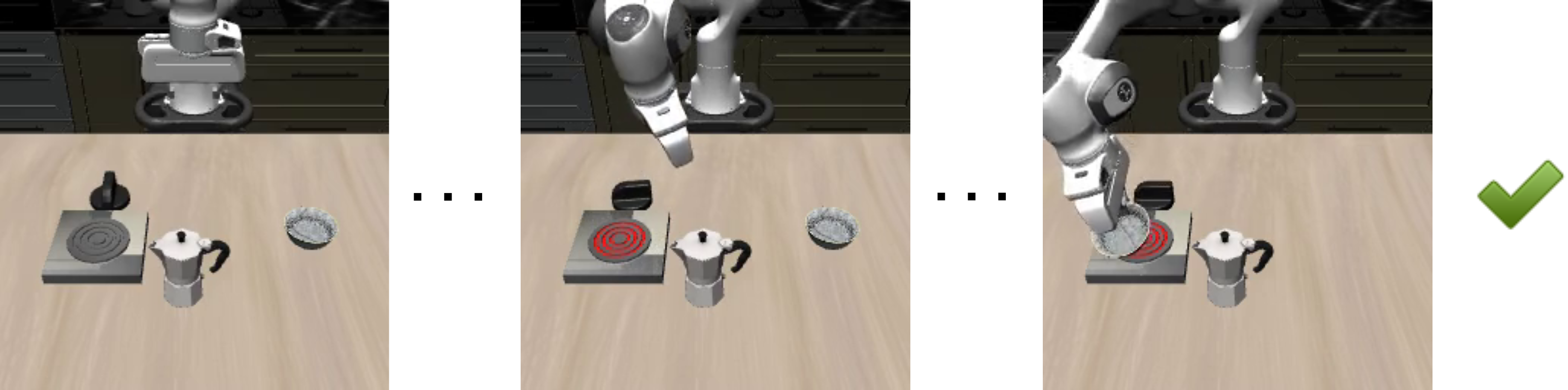}\\
\midrule
Pick up the chocolate pudding at the back and put it in the top drawer of the cabinet &
  \includegraphics[width=\linewidth]{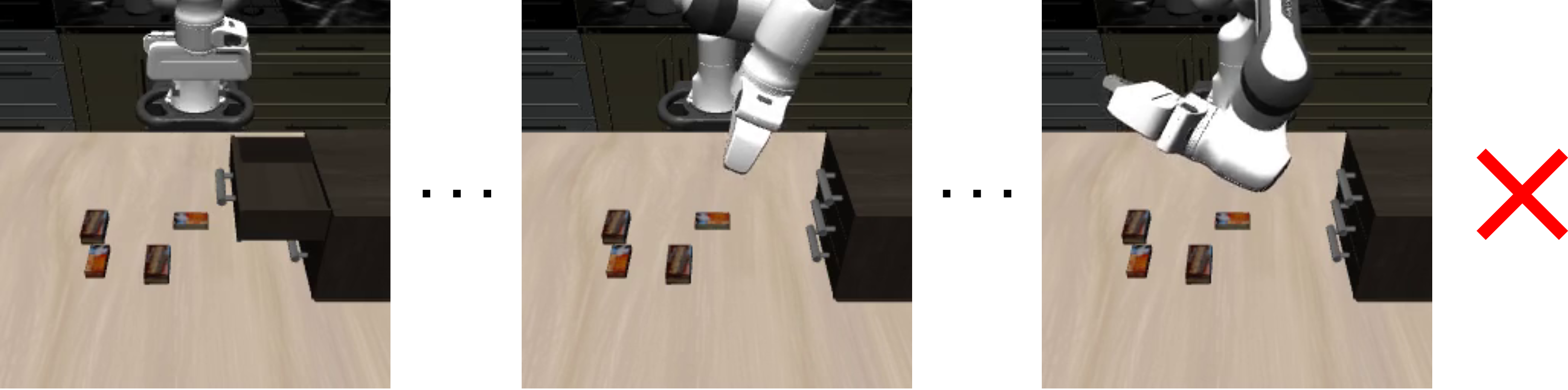}&
  \includegraphics[width=\linewidth]{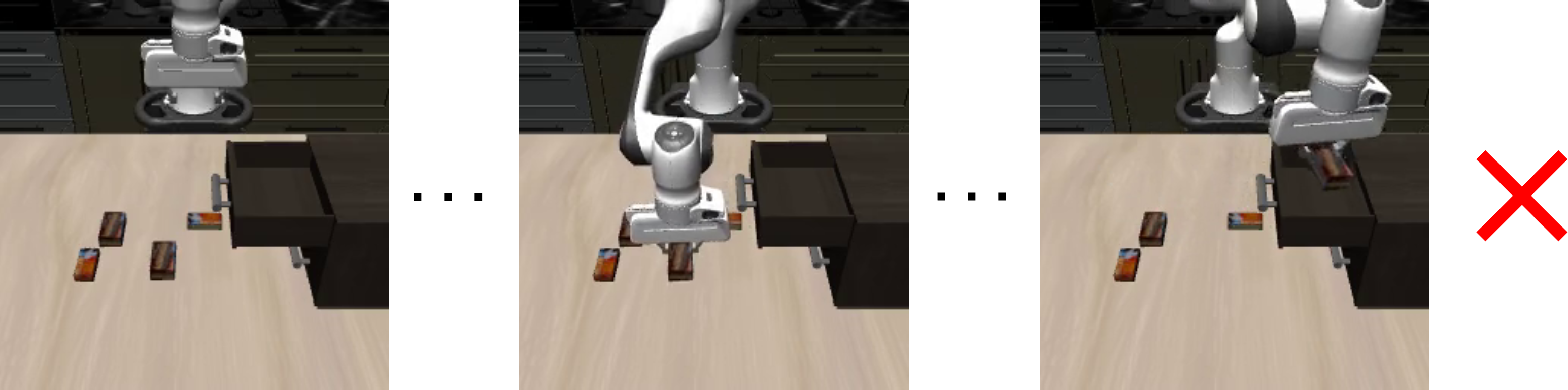}\\
  \midrule
Grab the white bowl and keep it on the stove &
  \includegraphics[width=\linewidth]{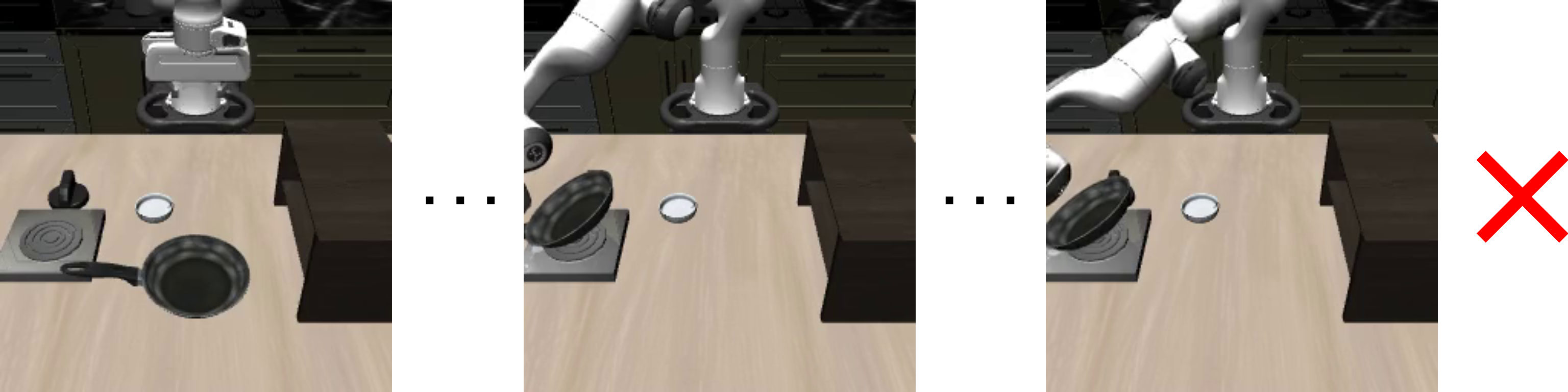}&
  \includegraphics[width=\linewidth]{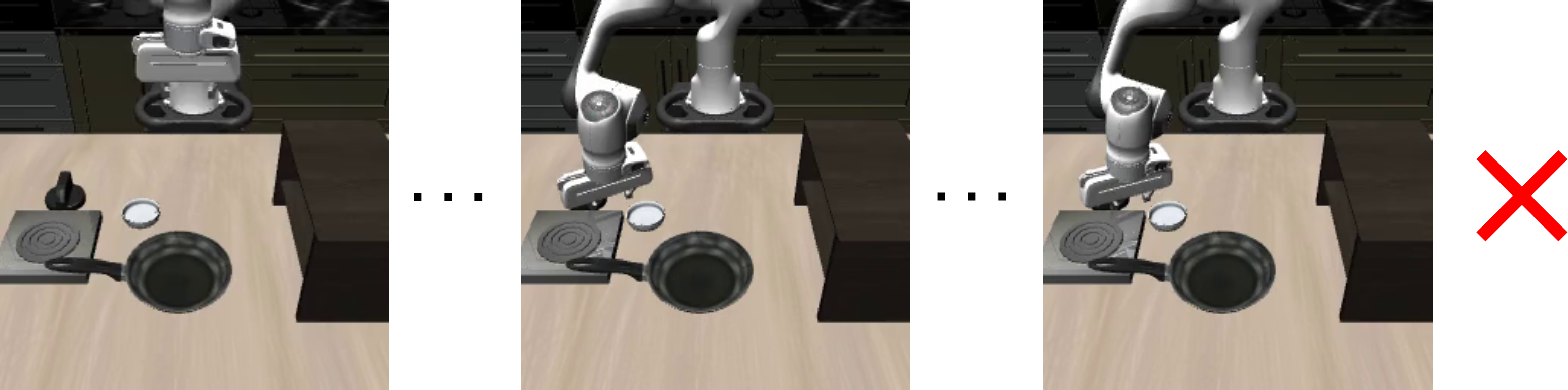}\\
\bottomrule[1pt]
\end{tabular}
\vspace{-10pt}
\label{tab:comparison_examples}
\end{table*}

\subsection{Zero-Shot Evaluation}


While the LIBERO suites test a variety of manipulation skills, models fine-tuned on these benchmarks tend to overfit, failing to generalize when task distributions shift. To assess zero-shot capabilities, we create the \emph{LIBERO-Unseen} benchmark by modifying BDDL specifications in the LIBERO-90 dataset. This new benchmark introduces 10 tasks where object identities and goal locations are familiar, but their combinations are novel. For example, a robot may have learned to grasp a white bowl and separately to place items on a stove, but has never been trained to execute both actions together as in ``Put the white bowl on the stove."  Each task is evaluated over 50 trials, with results reported in Table~\ref{tab:libero-unseen-results}. We also include an additional baseline ECoT*, which finetunes OpenVLA-OFT on embodied chain-of-thought reasoning from ECoT~\cite{c67}, to contrast predicting reasoning chains with using reasoning as input. 

3D-CAVLA outperforms ECoT* by 4.6\%, OpenVLA-OFT by 8.8\% and the Diffusion Transformer by 18.2\%.  The improvements stem from three factors: (i) chain-of-thought instructions allow decomposition of unseen tasks into sub-steps partially observed during training, (ii) ROI pooling focuses attention on relevant regions, reducing distractions, and (iii) depth features provide 3D cues that facilitate transferring skills to new contexts. These combined enhancements enable significantly more robust zero-shot performance. Although ECoT * produces robust reasoning traces in inference, these predictions remain tied to the distribution of training tasks. As a result, performance degrades on unseen tasks, where grounding tokens become less reliable, and on longer-horizon tasks, where errors compound over time. In contrast, 3D-CAVLA provides reasoning through frozen foundation models as policy input rather than learning to predict it within the VLA, which avoids this in-domain overfitting and yields more consistent performance under distribution shift. Furthermore, ECoT* requires either double the compute time by using separate parallel instances for reasoning and action prediction, or runs in its standard form at a lower control frequency ($\sim$1.5\,Hz). In contrast, OFT and 3D-CAVLA run at $\sim$4.4\,Hz on a single A100 GPU.

Example success and failure cases are illustrated in Table~\ref{tab:comparison_examples} where we compare rollouts between OpenVLA-OFT and 3D-CAVLA. Rows 2 and 3 highlight instances where 3D-CAVLA provides clear advantages. When faced with unseen tasks, OpenVLA-OFT often defaults to trajectories resembling those from similar training tasks, leading to incorrect executions. In contrast, 3D-CAVLA leverages ROI-guided visual restriction to reliably localize the target object, and chain-of-thought reasoning to generate motions consistent with task structure. Depth features further enhance precision, enabling the policy to place the mug and bowl accurately at the center of their respective receptacles. Integrating depth provides a significant gain in cluttered scenes, suggesting that explicit geometric reasoning is a key bottleneck in RGB-only VLAs. Direct point cloud encoding within policy learning enables better spatial disambiguation. 

The last two rows illustrate representative failure cases of 3D-CAVLA. In the first, errors arise from reliance on object masks generated by Molmo~\cite{c7}. When the VLM misidentifies an object, the resulting mask guides the policy toward an incorrect target, producing faulty motions. This highlights the need for stronger grounding, for example by incorporating geometric priors into the vision-language encoder to improve disambiguation of referring expressions. The final row shows a case where the white bowl is not grasped due to an unfavorable approach angle. Since the policy is primarily trained on grasps from a limited set of orientations, it struggles to generalize to collision-free strategies in novel contexts. These observations suggest that augmenting training with negative examples and explicit grasp diversity could improve robustness, allowing the model to learn not only from successful trajectories but also from failures that highlight unsafe actions.

\begin{figure*}[ht]
\vspace{5pt}
  \centering
   \includegraphics[width=\linewidth]{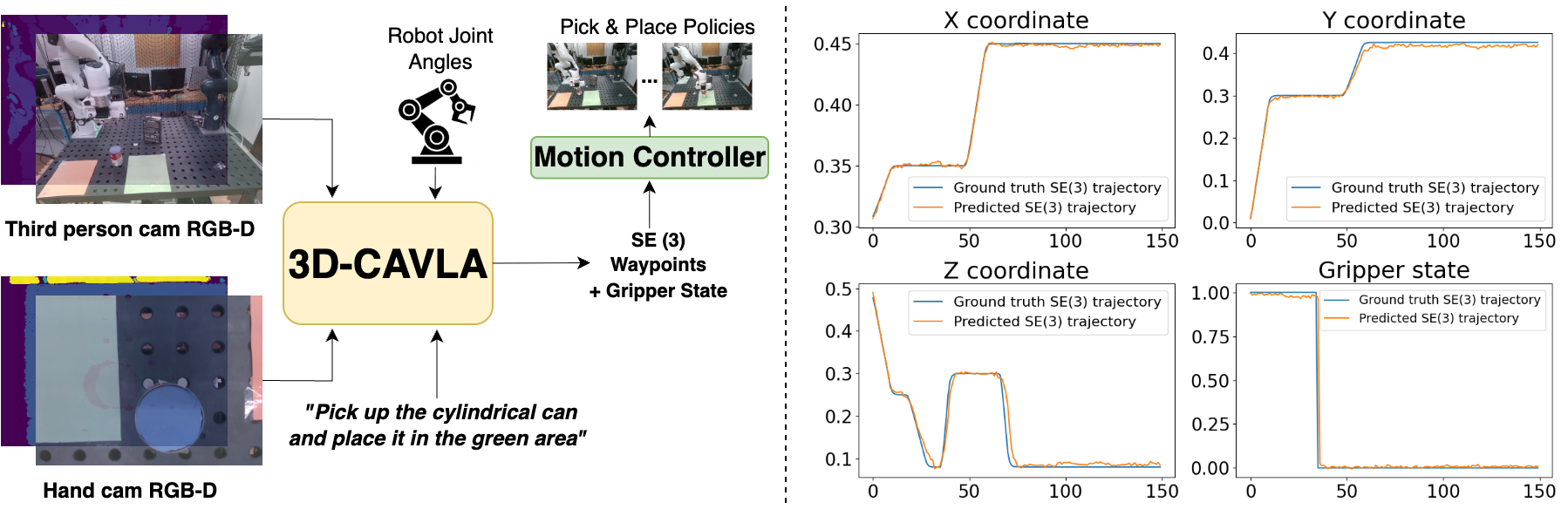}
   
   \caption{Real robot trials. 3D-CAVLA transforms vision-language observations to SE(3) waypoints and binary gripper states for task execution. We train policies in SE(3) and observe improved performance over velocities. 3D-CAVLA's waypoint policy closely matches ground truth (RHS) in successful trials.}
   \label{fig:real-robot-experiments}
   \vspace{-15pt}
\end{figure*}



\subsection{Ablation Studies}

Table~\ref{tab:ablations} reports ablations on key components of 3D-CAVLA. Removing depth features causes the sharpest decline, indicating that 3D perception is essential for robust manipulation. In unseen tasks, TA-ROI helps the policy focus on relevant regions of the RGB view to predict unseen motion, and textual expansion with CoT helps connect unseen tasks with seen subtasks. Removing TA-ROI pooling marginally improves performance on in-distribution tasks, as binary masks can exclude useful context (e.g., stationary drawer parts in ``open drawer and move bowl inside"). Ablation differences are more pronounced in unseen tasks, indicating that each component of 3D-CAVLA contributes primarily to generalization rather than in-distribution performance.

\begin{table}[!ht]
\caption{Ablation Studies. We remove key components of 3D-CAVLA and measure performance in both seen and unseen tasks.}
\centering
\setlength{\tabcolsep}{4pt} 
\begin{tabular}{lcc}
\toprule
\textbf{Method} & \textbf{Libero-Seen} & \textbf{Libero-Unseen} \\
\midrule
3D-CAVLA& 98.1 & 45.2  \\
\midrule
w/o CoT & 97.4 & 42.4 \\
w/o Depth & 97.0 & 41.0 \\
w/o TA-ROI & 98.2 & 41.4 \\
\bottomrule

\end{tabular}
\vspace{-10pt}
\label{tab:ablations}
\end{table} 

\begin{table}[ht]
\caption{Robot experiments evaluating Diffusion Policy, OpenVLA-OFT and 3D-CAVLA for seen, similar and unseen tasks; reported as success rate ($\%$) out of 10 trials per task.}
\centering
\small
\setlength{\tabcolsep}{3.0pt}
\begin{tabular}{l c c c}
\toprule
\textbf{Model} & \textbf{Seen} & \textbf{Similar} & \textbf{Unseen} \\
\midrule
Diffusion Policy & 84.2 & 46.0 & 21.8 \\
\midrule
OpenVLA-OFT & 88.6 & 54.4 & 30.2  \\
\midrule
3D-CAVLA & \textbf{90.0} & \textbf{60.0} & \textbf{38.0} \\
\bottomrule
\end{tabular}
\vspace{-15pt}
\label{tab:robot-exp}
\end{table}

\subsection{Real-World Robot Experiments}

\newcolumntype{U}{>{\columncolor{unseenyellow}}c}

To validate transfer to physical systems, we collect teleoperated demonstrations on a Franka robotic arm for ten tabletop pick-and-place tasks involving five objects (sugar box, soup container, mustard bottle, can of peas and soda can) and two target regions (orange, green). For each task, we record 50 demonstrations using an end-effector camera and an external third-person view (see Fig.~\ref{fig:real-robot-experiments}). We conduct five independent train-test evaluations, where policies are trained on seven random tasks (seen) and evaluate on the held-out three tasks (unseen) in ten trials per task. We also evaluate trained policies on compounded tasks, termed as ``similar" for example ``Place both objects in the green area". This tests reasoning and generalization to unseen text instructions with seen motions. Table~\ref{tab:robot-exp} summarizes the results. On seen tasks, both OpenVLA-OFT and 3D-CAVLA achieve comparable success rates. For similar tasks, all models struggle in performance. While the deterioration in Diffusion Policy and OpenVLA-OFT stems from overfitting to in-distribution text instructions, the CoT prompts in 3D-CAVLA helps break down the instruction into seen subtasks to improve performance. However a score of 60\% reflects the difficulties in long form action generation which is a key challenge for all methods. On unseen tasks, 3D-CAVLA consistently outperforms baselines, demonstrating that structured reasoning, 3D spatial awareness, and ROI-guided perception translate effectively from simulation to physical deployment. By restricting attention to task-relevant regions, TA-ROI accelerates training. For example, 3D-CAVLA converges in only 3K epochs—over \textbf{3$\times$ faster} than OpenVLA-OFT, which requires 10K epochs. 
Since both LLM prompted CoT and region of interest detection is computed only once before policy rollout, the test time frequency of both OpenVLA-OFT (4.4Hz) and 3D-CAVLA (4.3Hz) is similar.

\section{CONCLUSION AND FUTURE WORK}
\label{sec-conclusion}

We presented \textbf{3D-CAVLA}, a vision-language-action finetuning paradigm that incorporates chain-of-thought reasoning, depth-based point cloud features, and task-aware region-of-interest pooling. Together, these components transform the problem from purely 2D perception to a richer 3D understanding, enabling stronger reasoning and improved zero-shot generalization. Extensive experiments on the LIBERO benchmark and real-world Franka arm tasks demonstrate that 3D-CAVLA consistently outperforms competitive baselines, narrowing the gap between in-distribution success and robust deployment on unseen tasks. Our findings show that equipping VLAs with structured reasoning and 3D scene awareness is crucial for building generalizable robotic policies.  

Our results highlight the model’s capacity to generalize under practical conditions, but also reveal two key failure modes that limit performance.  First, in several trials, the policy reverted to executing trajectories resembling previously seen tasks, indicating overfitting to the relatively small dataset used for fine-tuning. Second, when approaching the target object, the robot frequently oscillated near the grasp point without completing the action, due to low variation in training images near contact and insufficient cues to trigger grasp closure.  We identify two directions to address these challenges: (i) Co-training with larger franka datasets from OXE~\cite{c33} would increase data diversity and mitigate overfitting, and (ii) Incorporating VLM guided reasoning from a static camera feed to provide useful textual feedback during execution and recovery from oscillatory states. Together, these enhancements could further narrow the gap between seen and unseen task performance in real-world settings.












\end{document}